# Compressed Inference for Probabilistic Sequential Models

**Gungor Polatkan**
Department of Electrical Engineering
Princeton University
Princeton, NJ 08544

**Oncel Tuzel**
Mitsubishi Electric Research Laboratories
Cambridge, MA 02139

## Abstract

Hidden Markov models (HMMs) and conditional random fields (CRFs) are two popular techniques for modeling sequential data. Inference algorithms designed over CRFs and HMMs allow estimation of the state sequence given the observations. In several applications, estimation of the state sequence is not the end goal; instead the goal is to compute some function of it. In such scenarios, estimating the state sequence by conventional inference techniques, followed by computing the functional mapping from the estimate is not necessarily optimal. A more formal approach is to directly infer the final outcome from the observations. In particular, we consider the specific instantiation of the problem where the goal is to find the state trajectories without exact transition points and derive a novel polynomial time inference algorithm that outperforms vanilla inference techniques. We show that this particular problem arises commonly in many disparate applications and present experiments on three of them: (1) Toy robot tracking; (2) Single stroke character recognition; (3) Handwritten word recognition.

## 1 Introduction

Assigning labels to sequential data is a common problem extensively studied in several application domains such as computer vision and computational linguistics (Rabiner (1989), Lafferty et al. (2001), Quattoni et al. (2004)). For instance, in part-of-speech tagging, the problem is to tag parts of speech by considering the grammatical structure of the language, e.g., (verb verb noun noun verb adjective) is a very unlikely sequence in English. Likewise, one can assign letters to a sequence of images of hand-written characters by again exploiting the structure enforced by the grammar of that language. In these examples, sequential patterns are important and they can be used to extract information from massive data sets.

Two common models for solving such problems are hidden Markov models (HMMs), and conditional random fields (CRFs, often as a linear-chain). These models have been extended in various forms to adapt to different types of problems. For example, semi-Markovian CRFs are introduced as a solution to segmentation problem allowing non-Markovian transitions in segments and assigning direct labels not to individual samples but to overall segments (Sarawagi & Cohen (2004)). Fox et al. (2008) proposes a non-parametric prior for systems with state persistence to prevent unrealistically many transitions. This method not only provides state persistence, but also allows learning the transition probabilities in an infinite state space.

In these examples, the inference algorithm estimates the state sequence. But in several applications, this is not the end goal; instead the goal is to compute some function of the state sequence. In particular, we consider a frequently occurring form of this problem, *compressed inference*, where the function $compress$ just keeps track of the state transitions without keeping track of the dwell times at each state and exact transition points.

A simple example is the detection of actions in the movement of a human subject where the exact transition points between states like "sitting($s$)", "jumping($j$)", "walking($w$)", and "running($r$)" are ambiguous and not important, but the detection of unique actions and the order of appearance is significantly important. For example, let the observation sequence be $\mathbf{x} = \{x_1, x_2, ....x_9\}$ and the corresponding true state sequence be $\mathbf{y} = \{s, s, j, j, j, w, w, r, r\}$. The end goal is the accurate prediction of the output of the function $compress(\mathbf{y})$ given the observation sequence $\mathbf{x}$, where $compress(\mathbf{y}) = \{s, j, w, r\}$ in this case. When a prediction $\mathbf{y}'$ is $\{s, s, j, j, j, j, w, r, r\}$ (which is acquired by converting a state '$w$' to '$j$' exactly at the transition from $j$ to $w$), it is an error for conventional applications, but it is not an error for this application, since $compress(\mathbf{y}) = compress(\mathbf{y}')$. Inversely, when a prediction $\mathbf{y}''$ is $\{s, s, j, j, w, j, w, r, r\}$, it is a fatal error for this application, even though it only differs from the ground truth sequence $\mathbf{y}$ by one state.

In contrast to standard sequence labeling problem, the length of the compressed output is also unknown, e.g., it is unclear how many unique actions occurred in the order of appearance during the movement of a human subject. More precisely, although $\mathbf{y}$ has the same length with $\mathbf{x}$, the length of $compress(\mathbf{y})$ is not known before hand, which might take values from 1 (means there are no state transitions) to the length of the sequence $\mathbf{x}$. Therefore the inference algorithm should estimate the length of the compressed sequence in conjunction with the states.

To the best of our knowledge, this is a problem largely unaddressed in machine learning. In this paper, we present a polynomial time algorithm to directly infer the length and the states of the compressed sequence using marginalization over all state sequences. The experiments show that the proposed inference algorithm consistently outperforms standard inference techniques using the same training model.

One important application domain that benefits from compressed inference is the labeling of sequence data. The conventional approach for sequence classification applications such as action recognition, gesture recognition, etc. requires training a separate sequence model for each action/event class. The classification task then assigns the new observations to the action/event classes according to the observation likelihoods among the trained models. This approach becomes unpractical or even infeasible when the number of action/event classes is very large such as recognizing all the words in a language. In such cases training a single sequence model and classifying the sequence data according to the unique states decoded is a feasible approach which can be obtained via compressed labeling.

# 2 Background: Conditional Random Fields and Inference Techniques

The sequence labeling problem can be formulated as finding the best function $f$ that can predict $\mathbf{y} = f(\mathbf{x})$, given $N$ training sequences $\{(\mathbf{x}_i, \mathbf{y}_i)\}_{i=1}^N$, where $\mathbf{x}_i = \langle x_{i,1}, x_{i,2}, \ldots.., x_{i,T_i}\rangle$ is the observation sequence and $\mathbf{y}_i = \langle y_{i,1}, y_{i,2}, \ldots.., y_{i,T_i}\rangle$ is the label sequence. Linear-chain CRFs and HMMs are two probabilistic models targeting this problem. Linear-chain CRFs can be thought as conditional HMMs, or HMMs can be thought as a special case of CRFs with a particular choice of feature function. While we present algorithms and results for CRFs, they are equally applicable to HMMs without loss of generality. We use linear-chain CRFs as our base learners throughout this paper. Next, we review CRFs and conventional inference techniques.

## 2.1 Conditional Random Fields(CRFs)

In a linear chain conditional random field of Lafferty et al. (2001), the conditional distribution is modeled as

$$p(\mathbf{y}|\mathbf{x}) = \frac{1}{Z(\mathbf{x})} \prod_{t=1}^{T} \Psi(y_t, y_{t-1}, x_t), \quad (1)$$

$$\Psi(y_t, y_{t-1}, x_t) = \exp\left(\sum_j \lambda_j g_j(y_{t-1}, y_t, x_t) + \sum_k \mu_k u_k(y_t, x_t)\right). \quad (2)$$

where $\Psi(y_t, y_{t-1}, x_t)$ is called the potential function; $g_j(y_{t-1}, y_t, x_t)$ is called the transition feature function from state $y_{t-1}$ to $y_t$; $u_k(y_t, x_t)$ is called the state feature function at state $y_t$; $\lambda_j$ and $\mu_k$ are the parameters estimated at the learning process, and $Z(\mathbf{x})$ is the normalization factor as a function of the observation sequence. In this paper, we assume that the trained model is given, and refer readers to Sutton & McCallum (2006) and Altun et al. (2003) for detailed discussions on learning model parameters.

## 2.2 Vanilla Inference Techniques

Here, we present a brief overview of conventional inference algorithms on probabilistic sequential models. One way of labeling a test sequence is the most likely labeling using the joint density $\mathbf{y}^* = \arg\max_{\mathbf{y}} p(\mathbf{y}|\mathbf{x})$. The solution can be efficiently computed via Viterbi algorithm using recursion $\delta_t(j) = \max_i \Psi(j, i, x_t)\delta_{t-1}(i)$, which propagates the most likely path based on the max product rule (Sutton & McCallum (2006)). However, in many applications, accurately predicting the whole label sequence is very difficult so that individual predictions are used. This is achieved via predicting $y_t$ from the marginal distribution $p(y_t|\mathbf{x})$ using a similar dynamic programming procedure, *forward-backward*. The forward recursion is given by $\alpha_t(j) = \sum_i \Psi(j, i, x_t)\alpha_{t-1}(i)$, where $\alpha_t(j)$ are the *forward variables*, and the backward recursion is given by $\beta_t(i) = \sum_j \Psi(j, i, x_{t+1})\beta_{t+1}(j)$, where $\beta_t(i)$ are the *backward variables*. The marginal probabilities can be computed by using these variables as given in Sutton & McCallum (2006).

In Culotta & McCallum (2004) and Kristjansson et al. (2004), a constrained forward algorithm is used to compute the confidence of a particular state sequence. The approach is to restrain the forward recursion to the constrained state sequence. In other words, given a set of constraints $Y' = \{y_q \ldots y_r\}$, a modified forward algorithm is used to compute the probability of any sequence satisfying $Y'$. The modified forward recursion is given as

$$\widehat{\alpha}_t(j) = \begin{cases} \sum_i \Psi(j, i, x_t)\widehat{\alpha}_{t-1}(i) & \text{for} \quad j \simeq y_{t+1} \\ 0 & \text{otherwise} \end{cases} \quad (3)$$

for all $y_{t+1} \in Y'$, where the operator $j \simeq y_{t+1}$ is defined in Culotta & McCallum (2004) as "$j$ conforms to constraint $y_{t+1}$". At time $T$, the confidence of a specific constraint is given as $Z'/Z$ where the constrained *lattice factor* $Z' = \sum_i \widehat{\alpha}_T(i)$ and the unconstrained *lattice factor* $Z = \sum_i \alpha_T(i)$ are computed by using constrained forward variables and unconstrained forward variables, respectively. When only a single constraint $y_t$ is included in the constraint set $Y'$, the method outputs the marginal distribution $p(y_t|\mathbf{x})$. Note that our algorithm for compressed inference is based on a similar idea.

## 3 Compressed Inference

In this section, we present the core of our inference algorithm designed to solve the compressed labeling problem. We first define a new sequence $\mathbf{s} = compress(\mathbf{y})$, e.g., if $\mathbf{y} = \{s, s, j, j, j, w, w, r, r\}$, then $\mathbf{s} = compress(\mathbf{y}) = \{s, j, w, r\}$. From now on, we use the symbol $\mathscr{C}$ to represent the function $compress$. The goal of compressed inference is to predict $\mathbf{s}$ given the observation $\mathbf{x}$. Throughout this paper, we use a traditionally trained Markov model (e.g. a linear chain CRF).

Next, we construct the mathematical framework for computing $p(\mathbf{s}|\mathbf{x})$. The overall joint density $p(\mathbf{s}|\mathbf{x})$ can be computed from $p(\mathbf{y}|\mathbf{x})$ by using the fundamental rules of probability theory.

**Proposition 3.1.** *Let* $\mathbf{y}$ *be a random variable taking values in E and* $\mathbf{s}$ *be a variable taking values in F. Then, the function* $\mathscr{C} : E \longmapsto F$ *is measurable and* $\mathbf{s}$ *is a random variable. Moreover, the conditional probability of* $\mathbf{s}$ *can be computed by the conditional probability of* $\mathbf{y}$ *as follows:*

$$p(\boldsymbol{s} = \boldsymbol{s}_0|\boldsymbol{x}) = \sum_{\forall \boldsymbol{y}:\mathscr{C}(\boldsymbol{y})=\boldsymbol{s}_0} p(\boldsymbol{y}|\boldsymbol{x}). \qquad (4)$$

*Proof.* A measure theoretic proof is given in the appendix. □

In a more verbal way, if one would like to compute $p(\mathbf{s} = \mathbf{s}_0|\mathbf{x})$, brute force approach is to find the set $Y' = \mathscr{C}^{-1}(\mathbf{s}_0)$, whose elements are all $\mathbf{y}$ sequences with compressed values $\mathbf{s}_0$, and then to compute the cumulative probability of $Y'$. Since we are working on discrete states, we just use summations in order to compute this probability.

This operation is a marginalization over all segmentations, $\mathbf{y}$, whose compressed values are $\mathbf{s}_0$. Unfortunately, given a model, though the computation of $p(\mathbf{y}|\mathbf{x})$ by using the methods in Section 2.2 is efficient, the final summation in equation (4) includes exponentially many operations ($M^T$, where $M$ is the number of states and $T$ is the length of $\mathbf{y}$), which is intractable. Next, we propose a novel polynomial time algorithm using dynamic programming.

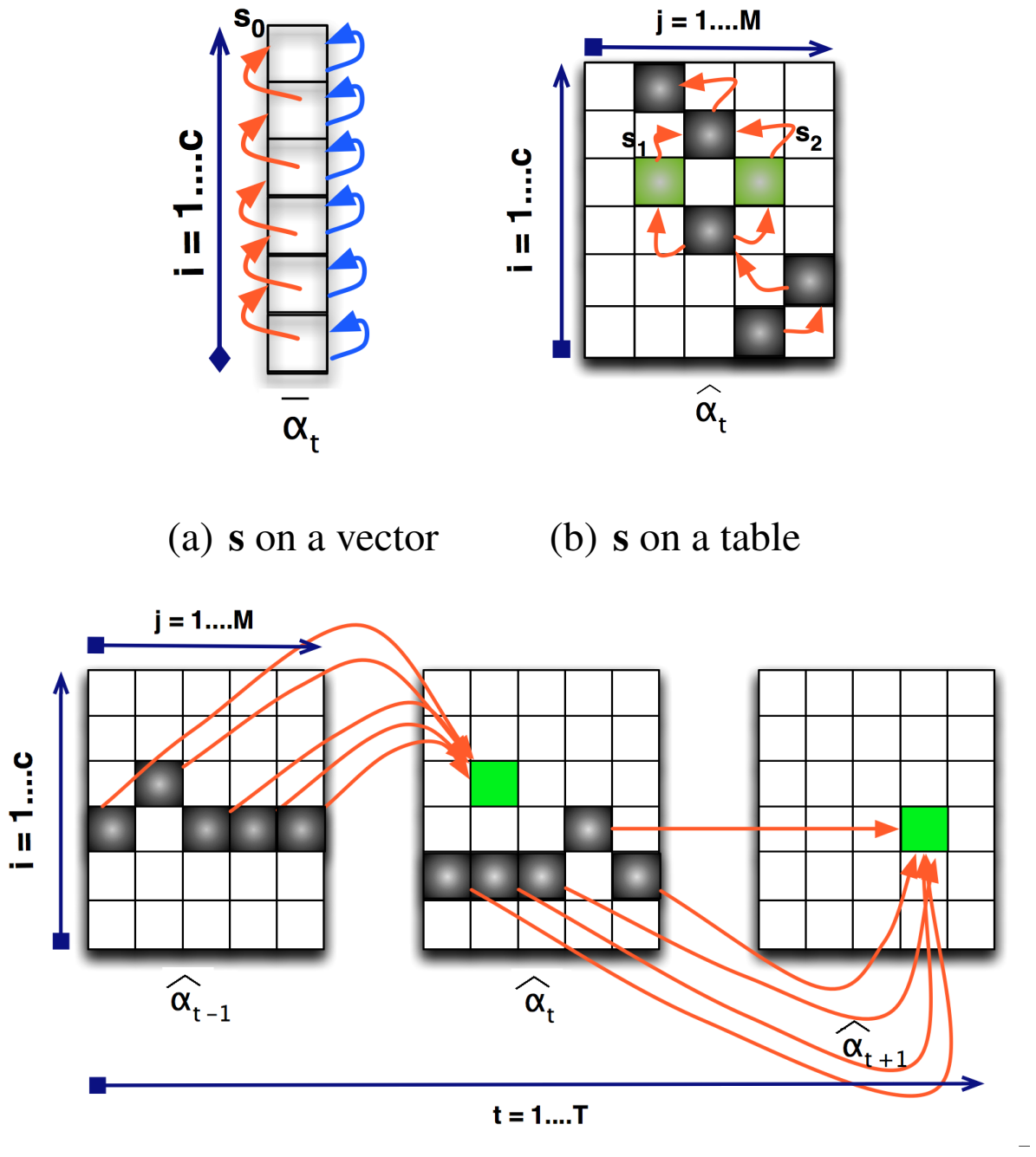

(a) **s** on a vector (b) **s** on a table

(c) Full two dimensional transitions

Figure 1: The recursions of compressed inference

### 3.1 Compressed Inference Algorithm

In this section, we first present solutions to three subproblems of compressed inference and then we perform compressed labeling using these techniques. Finally, we analyze the complexity of the proposed algorithm.

The first subproblem is computing the probability of a given compressed signal $\mathbf{s}_0$. In the second subproblem, we generalize this result and derive a dynamic programming algorithm to compute the probability distribution of the length of the compressed sequence. The third subproblem is computing the marginal probabilities of the compressed states. Finally, we perform labeling in the compressed domain by first finding the length of the compressed sequence from the distribution found in second subproblem, and then finding the compressed states from the marginal probabilities found in the third subproblem.

**Subproblem 1: Computing the probability of a compressed sequence $\mathbf{s}_0$, $p(\mathbf{s} = \mathbf{s}_0|\mathbf{x})$.** Here, we present a dynamic programming technique to compute this probability. This construction will help us in deriving the algorithm for the unknown $\mathbf{s}$ case which are explored by second and third subproblems.

Let $c = |\mathbf{s}_0|$ be the length of $\mathbf{s}_0$. For ease of notation, we refer to individual terms of $\mathbf{s}_0$ with $s_i, i = 1 \ldots c$. We define new forward variables $\overline{\alpha}_t(i), i = 1 \ldots c$, which keeps track of making exactly $i - 1$ transitions on sequence $\mathbf{s}_0$ up to time $t$. From time instance $t-1$ to $t$, the forward variables are updated based on: (1) Staying at the same state on observation $x_t$, which is shown with blue arrows in Figure 1(a); (2) Making a transition from $s_i$ to $s_{i+1}$, which is shown with orange arrows in Figure 1(a). The recursion is as follows:

**Proposition 3.2.** *The probability of a sequence* $\boldsymbol{s}_0, p(\boldsymbol{s} = \boldsymbol{s}_0|\boldsymbol{x})$*, is given by*

$$p(\boldsymbol{s}=\boldsymbol{s}_0|\boldsymbol{x}) = \sum_{\substack{\forall \boldsymbol{y}: \\ \mathscr{C}(\boldsymbol{y})=\boldsymbol{s}_0}} p(\boldsymbol{y}|\boldsymbol{x}) \propto \sum_{\substack{\forall \boldsymbol{y}: \\ \mathscr{C}(\boldsymbol{y})=\boldsymbol{s}_0}} \prod_{t=1}^{T} \Psi(y_t, y_{t-1}, x_t) \tag{5}$$

*which can be computed by the recursion*

$$\overline{\alpha}_t(i) = \Psi(s_i, s_{i-1}, x_t)\overline{\alpha}_{t-1}(i-1) + \Psi(s_i, s_i, x_t)\overline{\alpha}_{t-1}(i), \tag{6}$$

*where* $i = 1 \dots c$*. At time* $T$*, we attain*

$$\overline{\alpha}_T(c) = \sum_{\forall \boldsymbol{y}:\mathscr{C}(\boldsymbol{y})=\boldsymbol{s}_0} \prod_{t=1}^{T} \Psi(y_t, y_{t-1}, x_t). \tag{7}$$

*Proof.* See the appendix. □

By this recursion we compute the lattice factor $Z(\mathbf{s}_0) = \overline{\alpha}_T(c)$. We will explain the way to compute the overall normalization factor $Z$ later in this section which will convert $Z(\mathbf{s}_0)$ to a probability by $p(\mathbf{s}=\mathbf{s}_0|\mathbf{x}) = Z(\mathbf{s}_0)/Z$.

**Subproblem 2: Computing the probability distribution of the length of the compressed sequence, $p(c|\mathbf{x})$.** Given observations $\mathbf{x}$, the first step of finding the compressed labels $\mathbf{s}$ is to find the length $c$ of $\mathbf{s}$, where $c$ can take values from 1 (which means there is no state transition) up to the sequence length $T$ (which means there is a transition at every single time step). Note that, for all $c_0 > T$, $p(c = c_0|\mathbf{x})$ is trivially zero.

Let $\mathbb{S}_i$ be the set of all compressed state sequences of length $i$, i.e., $\mathbb{S}_i = \{\mathbf{s} : |\mathbf{s}| = i\}$, $i = 1...T$. It is obvious that $\mathbb{S}_i \bigcap \mathbb{S}_j = \emptyset$ for $i \neq j$. Then, the probability $p(c = c_0|\mathbf{x})$ can be written as

$$p(c=c_0|\mathbf{x}) = p(\mathbf{s}\in\mathbb{S}_{c_0}|\mathbf{x}) = \sum_{\substack{\forall \mathbf{s}': \\ |\mathbf{s}'|=c_0}} p(\mathbf{s}=\mathbf{s}'|\mathbf{x})$$

$$\propto \sum_{\forall \mathbf{s}':|\mathbf{s}'|=c_0} \sum_{\forall \mathbf{y}:\mathscr{C}(\mathbf{y})=\mathbf{s}'} \prod_{t=1}^{T} \Psi(y_t, y_{t-1}, x_t). \tag{8}$$

We first note that $p(\mathbf{s} = \mathbf{s}_0|\mathbf{x})$ gives the probability of one possible $\mathbf{s}_0$ of length $c$. Suppose we have two such signals: $\mathbf{s}_1$ and $\mathbf{s}_2$ as shown in Figure 1(b). Then, $p(\mathbf{s} = \mathbf{s}_1|\mathbf{x}) + p(\mathbf{s} = \mathbf{s}_2|\mathbf{x}) \propto \overline{\alpha}_T(c)_{\mathbf{s}_1} + \overline{\alpha}_T(c)_{\mathbf{s}_2}$, where $\overline{\alpha}_T(c)_{\mathbf{s}_i}$ means forward recursion was run for $\mathbf{s}_i$. However, these two signals are different at only one point in compressed domain. To be able to represent them on the same lattice and avoid multiple calculations, we extend the vector representation of $\overline{\alpha}_t$ into a table, as shown in Figure 1(b). We note that *conventional forward variable* $\alpha_t$ was $M$ dimensional, *previous* $\overline{\alpha}_t$ was $c$ dimensional, and *new* $\hat{\alpha}_t$ is $c \times M$ dimensional.

This new representation requires the signal to traverse the table through certain cells. From now on we call these cells "constraints". Let the set of all constraints on the lattice be $Q = \{..., q_{l-1}, q_l, q_{l+1}...\}$, where each constraint $q_l$ is a tuple of the coordinates of the nonzero entries on the table. For example, for a particular compressed sequence $\mathbf{s}_0$, it corresponds to $\{(1, s_1), (2, s_2) \dots (c, s_c)\}$, and for a particular set $\mathbb{S}_i$, it corresponds to all coordinates of the table with height $i$ which is denoted by $Q_{\mathbb{S}_i}$. The recursion for a given constraint set $Q$ is as follows:

$$\hat{\alpha}_t(i,j)_Q = \begin{cases} \begin{pmatrix} \Psi(j,j,x_t)\hat{\alpha}_{t-1}(i,j)_Q + \\ \sum_{\forall k:k\neq j}\big(\Psi(j,k,x_t) \\ \times\hat{\alpha}_{t-1}(i-1,k)_Q)\big) \end{pmatrix} & \text{if}(i,j)\in Q \\ 0 & \text{otherwise.} \end{cases} \tag{9}$$

This recursion propagates through all colored nodes (which correspond to nonzero entries) on the table in Figure 1(b) and ignores all empty nodes since they are not included in $Q$. For simplicity, self loops are removed from Figure 1(b) . Moreover, the recursion for which the constraints included all the locations of the table (all entries on the table can be visited) is explained schematically in Figure 1(c) from $t-1$ to $t$ and from $t$ to $t+1$.

The recursion given in equation (9) computes the probability of all compressed sequences which are defined by the set $Q$ via

$$p(Q|\mathbf{x}) \propto Z(Q) = \sum_j \hat{\alpha}_T(c_0, j)_Q.\text{[1]} \tag{10}$$

Using constraint set notation $Q_{\mathbb{S}_{c_0}}$, the probability of $p(c = c_0|\mathbf{x})$ can be written as

$$p(c=c_0|\mathbf{x}) \propto \sum_{\substack{\forall \mathbf{s}: \\ |\mathbf{s}|=c_0}} \sum_{\substack{\forall \mathbf{y}: \\ \mathscr{C}(\mathbf{y})=\mathbf{s}}} \prod_{t=1}^{T} \Psi(y_t, y_{t-1}, x_t)$$

$$= Z(Q_{\mathbb{S}_{c_0}}) = \sum_j \hat{\alpha}_T(c_0, j)_{Q_{\mathbb{S}_{c_0}}}. \tag{11}$$

This corresponds to running the recursion in equation (9) with constraint set $Q_{\mathbb{S}_{c_0}}$ and summing the entries at row $c_0$. As we discussed before, $p(c = c_0|\mathbf{x}) = 0$ when $c_0 > T$ or

[1] The proof of this proportionality is just a generalization of the proof of Proposition (3.2) as given in the supplementary material.

$c_0 < 1$. If we employ this procedure for the constraint set $Q_{\mathbb{S}_T}$, the row sums of the table $\hat{\alpha}_T(i,j)_{Q_{\mathbb{S}_T}}$ produces all the lattice factors $Z(Q_{\mathbb{S}_i}), i = 1 \ldots T$ simultaneously. The overall summation of this table is equal to the normalizing factor $Z$ which is necessary for computing $p(\mathbf{s} = \mathbf{s}_0|\mathbf{x}) = Z(\mathbf{s}_0)/Z$ and $p(c = c_0|\mathbf{x}) = Z(Q_{\mathbb{S}_{c_0}})/Z$. This identity follows from the fact that $Z$ is equal to the summation of the lattice factors for all possible lengths and combinations of $\mathbf{s}$.

**Subproblem 3: Computing the marginal probabilities of the compressed state sequences, $p(s_i = j|\mathbf{x}, c)$.** To compute the marginal distribution $p(s_i = j|\mathbf{x}, c)$, we construct the constraint set $Q_{i,j}$ by including all the entries of the table with height $c$ except the ones at row $i$. Then we add $(i, j)$ to this set. This particular constraint set configuration includes all the possible compressed sequence configurations with length $c$ and $s_i = j$. Then, the marginal probability is computed by $p(s_i = j|\mathbf{x}, c) = Z(Q_{i,j})/\sum_j Z(Q_{i,j})$.

**Compressed labeling.** The compressed labeling is a simple application of the aforementioned methods: (1) Estimate $c$ by $\hat{c} = \arg\max_{c_0} p(c = c_0|\mathbf{x})$; (2) Estimate $s_i$ by $s_i = \arg\max_j p(s_i = j|\mathbf{x}, \hat{c})$.

**Computational Complexity.** The time complexity of the Viterbi algorithm is $O(TM^2)$ and space complexity is $O(TM)$ where $T$ is the length of the sequence and $M$ is the number of states. The new recursion propagates on a two dimensional table which requires $O(c_{max}TM^2)$ time and $O(c_{max}TM)$ space where $c_{max}$ is the maximum feasible length of the compressed sequence. Although the maximum possible value of $c_{max}$ is equal to $T$, in general $c_{max} \ll T$.

## 4 Experiments

Experiments are performed on three different applications: toy robot tracking, single stroke character recognition and handwritten word recognition. In all experiments, we use both marginal (using forward-backward) and joint posterior (using Viterbi) estimates as input to the compress function $\mathscr{C}$ to obtain $\mathbf{s}_{marginal}$ and $\mathbf{s}_{joint}$ as final decisions for the baseline predictions of $\mathbf{s}$. We compare these baseline predictions with the proposed compressed inference algorithm described in Section 3.1. We used the hCRF package (Morency (2007)) with the BFGS optimizer to learn the CRF model parameters. In addition, we present comparison with the semi-Markov CRF model in the first two problems where the training and inference are performed using the semi-Markov CRF package (Sarawagi (2009)). The semi-Markov CRF model produces segmentation of the input sequence where we discard the transition points and the labels of the segments give the compressed sequence.

**Metrics.** We consider two evaluation metrics which we call *Exact* score and Edit Distance Score *(EDS)*. *Exact* score is used in order to check the perfect match of the compressed predictions to the truth. If we miss even one state in $\mathbf{s}$ or $\mathscr{C}(\mathbf{y})$, this is a fatal error and we consider this sequence as missed during the evaluation. Hence, both the compressed length of the prediction and exact compressed state sequence should perfectly match with the truth. In *EDS*, we measure how well we are performing state by state. We use *edit distance* between two strings of characters which is defined as the minimum number of operations required to convert one to the other. While computing *EDS*, in order to prevent the effect of the length of the sequences, we normalize EditDistance(prediction, truth) of each sequence by $\max_{Length}\{$prediction, truth$\}$. Then, we sum the normalized distances of all sequences and finally we find *EDS* $= 100 - 100 *$ sum$/(\#$ of sequences$)$.

**Toy Robot Tracking.** Sequential models are frequently used in robot tracking applications, where a robot is moving in a small grid-based environment to discover the world (Baltzakis & Trahanias (2002)). If one is interested in finding the visited locations only, the unique grids that the robot goes through is important but the self-loops and the exact transition times/boundaries are not important. For instance, our robot moves in a small world as seen in Figure 2(a), where transition possible grids are {blue, green, yellow, red} colored and obstacles, which prevent motion are {black}. At every step, the robot attempts to move {up, down, left, right} based on a random choice of direction. If there is a block towards the intended direction, it tries again. In this problem, state refers to the location of the robot in $(x{:}y)$ coordinates, and observation refers to an observed color which is the output of a sensor (a camera for instance) with a color detection accuracy of $P\%$.

We generated 400 training and 400 test sequences at each of 6 different $P$ values ({100, 90, 80, 70, 60, 50}) by simulating the robot in the small world as defined in Figure 2(a). The sequence lengths are in the range of 100 - 300 and the compressed lengths are in the range of 5 - 15. In terms of exact score, our algorithm introduces an improvement when noise level is high or moderate (Figure 2(c)). But when noise level is very low, we observe that joint estimate performs better. We note that for very long state sequences (as used in this experiment), the exact score is not very reliable since a single state error causes a sequence to be labeled incorrect which drastically changes final outcome. In terms of EDS, as we can see in Figure 2(d), our state inference algorithm performs better than vanilla inference algorithms at high and moderate noise and at very low noise performance is similar.

The predictions of the semi-Markov CRF model are significantly inferior compared to the proposed compressed inference algorithm: the EDS based scores are $\%\{45.4,$ 40.7, 32.2, 15.5, 5, 4.5 } where as the Exact scores are %{10.5, 3.25, 2, 1, 0, 0 } for $P$ values of {100, 90, 80, 70, 60, 50} respectively. We attribute the poor performance of this model to the large number of maximum segmentation lengths. Likewise, the training and the inference of semi-Markov CRF models are extremely slow compared to the original CRF model (and compressed inference). The training takes 100 hours compared to 1-2 hours for the original CRF model. Please see Section 5 for a more detailed discussion.

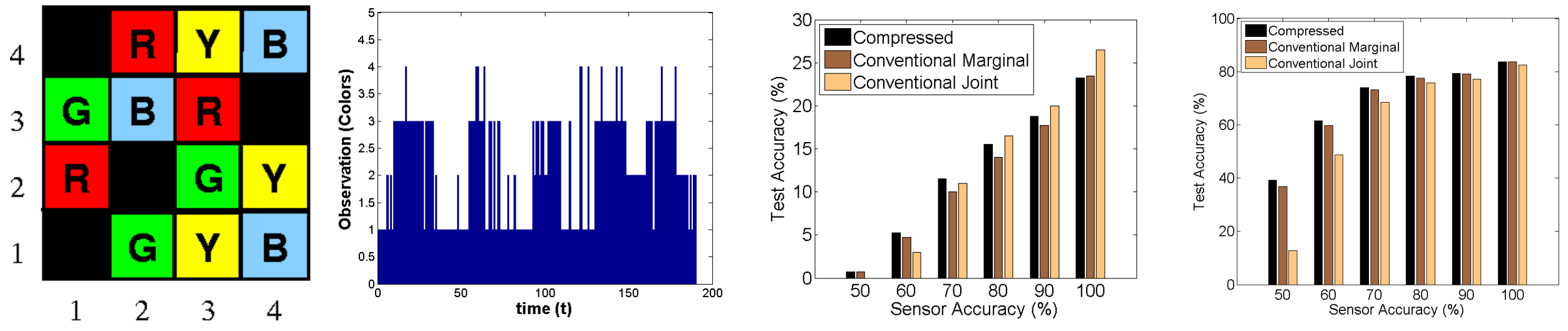

(a) Small World (b) Observations (c) Exact (d) EDS

Figure 2: a) Small world with coordinates and true colors of each square, b) A sample sequence of the robot in the small world in time domain representation. Each color is coded with a number. {b, g, y, r} with {1,2,3,4} respectively, c) Exact score d) EDS.

**Character Recognition.** In this experiment, we apply compressed inference to single stroke character recognition application (Ozun et al. (2001)). The problem is to recognize the shape drawn on a touch screen. It is assumed that the drawing operation is performed by a single stroke. One well known alphabet which has this property is the Graffiti$^{\text{TM}}$.

In this application, state refers to directions {up,right,down,left} and observation refers to quantized angles between successive points acquired from the user interface as shown in Figure 3(a) and 3(b). Existing systems use stochastic finite state machines (FSM) or HMMs for this purpose (Ozun et al. (2001)). Usually, an HMM or FSM is trained for each single character, then, one class is chosen by a likelihood test. One drawback of this method is its limited capability of handling arbitrary shapes. One can train a single model for all characters and decode the states by using this single model as well. Nevertheless, a single state error can spoil the whole prediction in such a setting. Hence, we need a strong inference scheme which is robust to noise. Moreover, the ambiguity in state transitions is an important obstacle as well, since passing from one state to another is generally ambiguous. Our approach is robust to all these issues since it does not spend effort to estimate the exact transition locations but only produces transition sequences which is sufficient for this task.

We generated a data set of 20 training and 20 test samples for each character from the set {0,1,2,3,4,5,6,7,8,9} by using a computer user interface as shown in Figure 3(a). As shown in Table 3(d), compressed inference outperforms conventional inference. It especially performs better on characters like 0 and 3 which are the ones with the most ambiguous state transitions.

The results of the semi-CRFs match that of the joint estimate, %60 for the exact and %87 for the EDS based score, whereas the compressed inference algorithm significantly outperforms both, %95.5 and %98.97 respectively. Particularly, this problem is a good example where the transition boundaries are ambiguous. The semi-Markov CRFs force a segmentation over ambiguous boundaries whereas our model benefits from marginalization over all possible segmentations.

**Handwritten Word Recognition.** In this application, we use the data set in Taskar et al. (2003), which includes $16 \times 8$ size characters from the English alphabet. In the literature, handwritten word recognition is generally performed by first segmenting the characters and then recognizing them by multi class classification such as SVMs. In many studies, the structure of language is used as well (Taskar et al. (2003)). However, in all these works, the characters are already segmented during pre-processing step. Our setup is more challenging compared to these studies, i.e., the characters are not segmented and as a result the lengths of the sequences are not known as well.

For experimental purposes, we specified 20 names used in English {jake, conor, taner, wyat, cody, dustin, luke, jack, scot, logan, deshawn, deandre, marquis, darnel, terel, malik, reginald, tyrone, wilie, dominique} where none of them including any successive letters such as ‘nn’, etc. (Application can be generalized to arbitrary number of words and currently we do not consider words with successive letters which might be handled specially). Each word is written by different combinations of styles, 20 times for training set and 20 times for test set. Moreover, each time step corresponds to one vertical column of the image of handwriting,

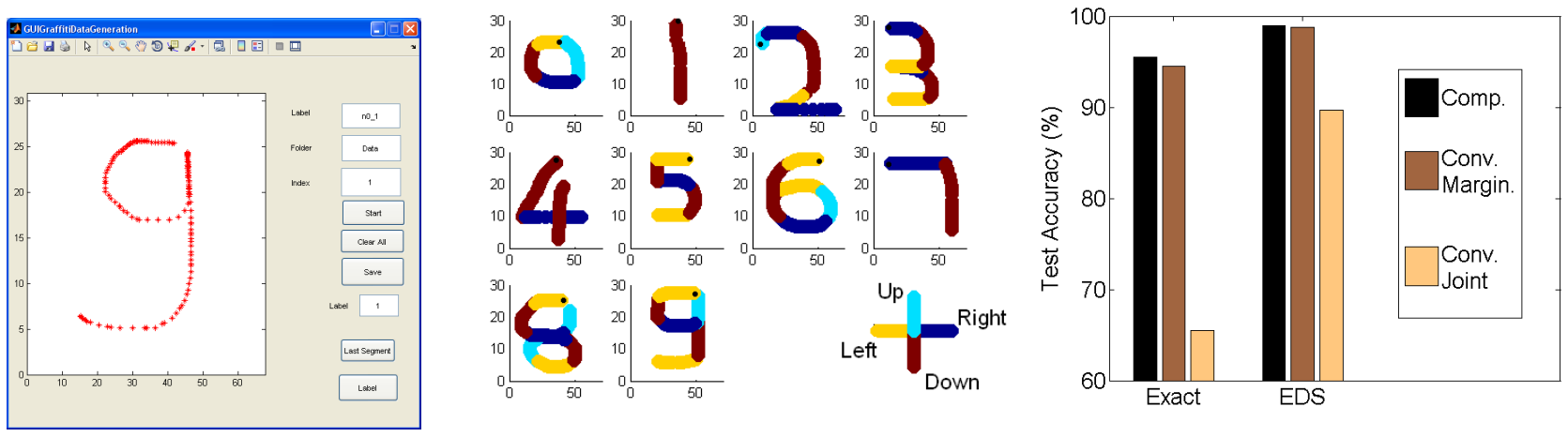

(a) GUI (b) A set of characters (c) Test Results

| | Exact | EDS |
|---|---|---|
| Comp. | **95.5** | **98.97** |
| Mar. | 94.5 | 98.75 |
| Joint | 65 | 89.67 |

d) % Scores

Figure 3: a) GUI for data generation, b) A set of characters with state labels; Results in c) graphical form, d)tabular form, in terms of EDS and Exact scores.

states refer to the corresponding character of the time step, and observations correspond to the shape context features of Belongie et al. (2001).

For feature extraction, we first take overlapping $16 \times 7$ patches from the word image by sliding window technique. Next, we apply shape context descriptor of Belongie et al. (2001) for each data point on the corresponding patch as shown in Figure 4(d). Then, we apply K-means clustering to learn a dictionary. In our experiments, $K = 50$ produced the best performance. Next, we generate histograms via vector quantization, which completes the feature extraction.

In the experimental results, we observe improvements in both scores compared to vanilla inference techniques, as can be seen in Figure 4(c) and Table 4(e).

## 5 Related Work

Compressed labeling of sequences is mentioned as a video-interpretation application in Fern & Givan (2004), where we get the function name "*compress*" from. Their main focus was to model a sequence problem that had an enormous number of states which was not known before hand. Due to the unknown number of states the conventional probabilistic models can not be used, hence the approach is applicable to a limited domain. In contrast our inference scheme uses a standard CRF or HMM and we do not make any assumption other than usual sequential modeling assumptions. Therefore the presented algorithm is applicable to any problem where distinct states are important and state transitions are ambiguous.

A simple transition-cost model is proposed in Fern (2005) for video interpretation where self transition is assumed to have no cost whereas all other possible transitions are assumed to have the same cost $K$. This is similar to training a probabilistic sequential model which has zero weights for all self transition parameters and same $K$ as the weight for all other transitions. However, this ad-hoc assumption is unrealistic for many applications.

Segmentation is the process of identifying the boundaries between segments (e.g., words in natural language processing, set of pixels in image processing). The output of segmentation is a set of segments with exact boundary locations (e.g. super-pixels with contours in images or state trajectories with exact state transition points in sequential models). The Semi-Markovian approach of Sarawagi & Cohen (2004) proposes a solution to the segmentation problem. Semi-Markov models explicitly model duration in a state with different distributions (which violates the Markovian assumption) and are widely used when one is interested in exact transition points and segmentation boundaries. In contrast to Sarawagi & Cohen (2004), our inference algorithm is designed for Markov models and benefits from ambiguities in segmentation boundaries via marginalizing over all boundary locations.

The problems where semi-Markov CRF models were shown to be successful, such as name entity recognition (Sarawagi & Cohen (2004)), have relatively short maximum segmentation lengths (around 3-4). In addition, in such problems the segmentation boundaries are very well defined. The experiments in this paper show that when the maximum segmentation length is large (in most of our applications it ranges from 50 to 100) or the transition boundaries are ambiguous, the compressed inference algorithm significantly outperforms semi-Markov CRFs. Moreover, the training time of the semi-Markov model is also prohibitive when the maximum segmentation length is large, which is another motivation for using a Markov model.

## 6 Discussion

Maximum likelihood or marginal estimate of a full state sequence is the standard approach for inference on Markov models. In this paper, we have shown that when the problem is finding the state trajectories only, without exact transition points, inference can be done in a more accurate way. To directly infer the unique states, we have proposed marginalization over possible transitions and derived a polynomial time algorithm. In three different applica-

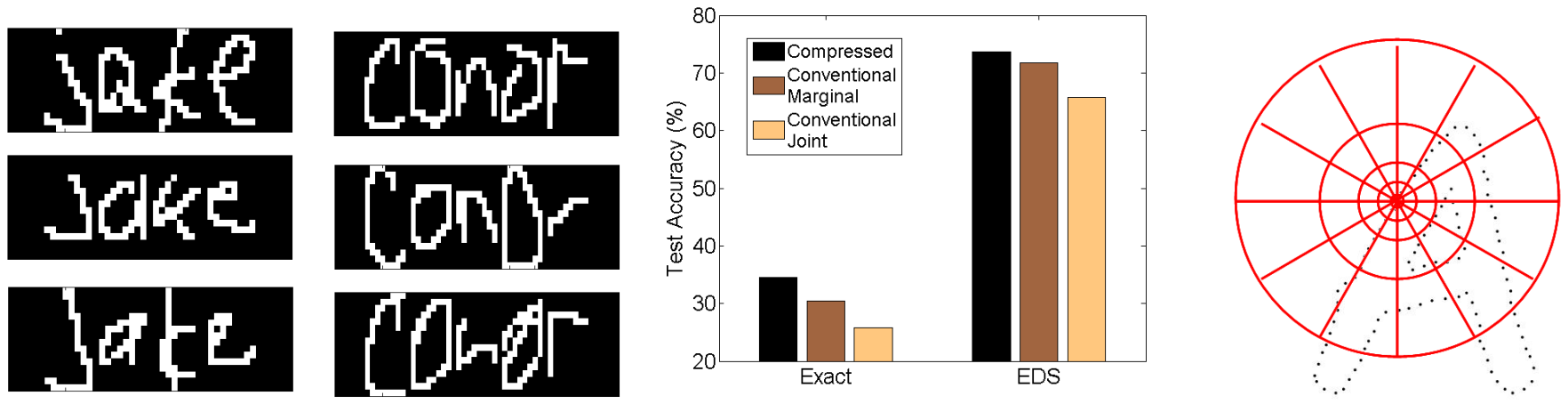

(a) Word *jake* (b) Word *conor* (c) Test Results (d) Shape Context

| | Exact | EDS |
|---|---|---|
| Comp. | **34.5** | **73.65** |
| Mar. | 30.5 | 71.65 |
| Joint | 25.75 | 65.72 |

e) % Scores

Figure 4: 3 instances of a) *jake*, b) *conor*; c) Results; d) Shape context feature extraction of Belongie et al. (2001); e) Results in terms of EDS and Exact scores.

tion domains, we have shown that the proposed compressed labeling algorithm outperforms vanilla techniques particularly when there is state transition ambiguity.

The proposed construction also offers significant potential for future research. Here, we have proposed polynomial time inference using marginal probabilities given the prediction for the length of the signal. Inference of the most likely joint compressed sequence is still an open problem which is unlikely to be polynomial time computable.

## References

Altun, Y., Johnson, M. & Hofmann, T. (2003), Investigating Loss Functions and Optimization Methods for Discriminative Learning of Label Sequences, *in* 'Proc. EMNLP'.

Baltzakis, H. & Trahanias, P. (2002), Hybrid mobile robot localization using switching state-space models, *in* 'ICRA', pp. 366–373.

Belongie, S., Malik, J. & Puzicha, J. (2001), 'Shape matching and object recognition using shape contexts', *IEEE Transactions on Pattern Analysis and Machine Intelligence* **24**, 509–522.

Culotta, A. & McCallum, A. (2004), Confidence estimation for information extraction, *in* 'Human Language Technology Conference and North American Chapter of the Association for Computational Linguistics (HLT-NAACL)'.

Fern, A. (2005), A simple-transition model for relational sequences, *in* 'IJCAI', pp. 696–701.

Fern, A. & Givan, R. (2004), Relational sequential inference with reliable observations, *in* 'Proc. of the International Conference on Machine Learning'.

Fox, E. B., Sudderth, E. B., Jordan, M. I. & Willsky, A. S. (2008), An hdp-hmm for systems with state persistence, *in* 'Advances in Neural Information Processing Systems'.

Kristjansson, T., Culotta, A. & Viola, P. (2004), Interactive information extraction with constrained conditional random fields, *in* 'AAAI', pp. 412–418.

Lafferty, J., McCallum, A. & Pereira, F. (2001), Conditional random fields: Probabilistic models for segmenting and labeling sequence data, *in* 'International Conference on Machine Learning', Morgan Kaufmann, pp. 282–289.

Morency, L.-P. (2007), 'HCRF library: Discriminative models for sequence labeling', *http://sourceforge.net/projects/hcrf/* .

Ozun, O., Ozer, O. F., Tuzel, C. O., Atalay, V. & Cetin, A. E. (2001), 'Vision based single stroke character recognition for wearable computing', *IEEE Intelligent Systems and Applications* **16**, 33–37.

Quattoni, A., Collins, M. & Darrell, T. (2004), Conditional random fields for object recognition, *in* 'Advances in Neural Information Processing Systems', MIT Press, pp. 1097–1104.

Rabiner, L. R. (1989), A tutorial on hidden markov models and selected applications in speech recognition, *in* 'Proceedings of the IEEE', pp. 257–286.

Sarawagi, S. (2009), 'semi-CRF :a java implementation of conditional random fields for sequential labeling', *http://crf.sourceforge.net/* .

Sarawagi, S. & Cohen, W. W. (2004), Semi-markov conditional random fields for information extraction, *in* 'Advances in Neural Information Processing Systems', pp. 1185–1192.

Sutton, C. & McCallum, A. (2006), *Introduction to Conditional Random Fields for Relational Learning*, MIT Press.

Taskar, B., Guestrin, C. & Koller, D. (2003), Max-margin markov networks, *in* 'Neural Information Processing Systems Conference', MIT Press.

# A Appendix

### Proof of Proposition 3.1

*Proof.* Let $(\Omega, \mathcal{H}, \mathbb{P})$ be a probability space where $\Omega$ is a set, $\mathcal{H}$ ( whose elements are called the *events*) is a $\sigma-$algebra on $\Omega$, and $\mathbb{P}$ is a probability measure on $(\Omega, \mathcal{H})$. Let $(E, \mathfrak{E})$ and $(F, \mathfrak{F})$ be measurable spaces, where $E$ and $F$ are two sets, and $\mathfrak{E}$ and $\mathfrak{F}$ are $\sigma-$algebras on $E$ and $F$, respectively. Let $\mathbf{y}$ be a random variable taking values in $(E, \mathfrak{E})$. Next, we analyse the measurability of $\mathscr{C}$, which will be useful while showing that $\mathscr{C}(\mathbf{y})$ is a random variable. Next, we define the inverse of a mapping: [1]

**Definition** A function $h$ from $E$ to $F$, i.e. $h : E \longmapsto F$, is a mapping which takes each $\mathbf{y}$ in $E$ and assigns to an element $h(\mathbf{y})$ in $F$. For any subset $A$ in $F$, the inverse image of $A$ under $h$ is defined as

$$h^{-1}A = \{y \in E : h(\mathbf{y}) \in A\} \tag{12}$$

**Proposition A.1.** *The function $\mathscr{C} : E \longmapsto F$ is measurable relative to $\mathfrak{E}$ and $\mathfrak{F}$.*

*Proof.* For every $A \in \mathfrak{F}$, $\mathscr{C}^{-1}A$ is the set of all $\mathbf{y}$ whose compressed values are in $A$, where $A = \bigcup_i \{\mathbf{s} = \mathbf{s}_i\}$. That is to say,

$$\begin{aligned}\mathscr{C}^{-1}A &= \mathscr{C}^{-1} \bigcup_i \{\mathbf{s} = \mathbf{s}_i\} = \bigcup_i \mathscr{C}^{-1}\{\mathbf{s} = \mathbf{s}_i\} \\ &= \bigcup_i \{\mathbf{y} \in E : \mathscr{C}(\mathbf{y}) = \mathbf{s}_i\}.\end{aligned}$$

Since $\{\mathbf{y} \in E : \mathscr{C}(\mathbf{y}) = \mathbf{s}_i\} \in \mathfrak{E}$ for all $i$, it is obvious that $\bigcup_i \{\mathbf{y} \in E : \mathscr{C}(\mathbf{y}) = \mathbf{s}_i\} \in \mathfrak{E}$, because a $\sigma$-algebra is closed under union. As a result, $\mathscr{C}^{-1}A \in \mathfrak{E}$ for every A in $\mathfrak{F}$ which completes the proof by definition of measurability. □

**Proposition A.2.** *The variable defined by $\mathbf{s} = \mathscr{C}(\mathbf{y})$, that is to say,*

$$\mathbf{s}(\omega) = \mathscr{C} \circ \mathbf{y}(\omega) = \mathscr{C}(\mathbf{y}(w)), \text{ where } \omega \in \Omega, \tag{13}$$

*is a random variable taking values in $(F, \mathfrak{F})$.*

*Proof.* By Proposition (A.1), $\mathscr{C}$ is measurable. Since measurable functions of measurable functions ($\mathbf{y}$ is a random variable) are measurable, we are done. □

Now, we know that $\mathbf{y}$ and $\mathbf{s}$ are random variables. Next question is how to find the distribution of $\mathbf{s}$. We observe that if $\phi$ is the distribution of $\mathbf{y}$, then the distribution $\kappa$ of $\mathbf{s}$ is $\kappa = \phi \circ \mathscr{C}^{-1}$, in other words,

$$\kappa(A) = \mathbb{P}\{\mathbf{s} \in A\} = \mathbb{P}\{\mathbf{y} \in \mathscr{C}^{-1}A\} = \phi(\mathscr{C}^{-1}A),\ A \in \mathfrak{F},$$

where $\phi(\mathscr{C}^{-1}A)$ corresponds to taking the integral of the set $\mathscr{C}^{-1}A$ with respect to the measure $\phi$. In a more verbal way, if one would like to compute $p(\mathbf{s} = \mathbf{s}_0|\mathbf{x})$, where $\mathbf{s}_0$ is in $A$, brute force approach is to find the set $Y' = \mathscr{C}^{-1}(\mathbf{s}_0)$, whose elements are all $\mathbf{y}$ sequences with compressed values $\mathbf{s}_0$, and then to take the integral with respect to $\phi$. Since we are working on discrete states, integrals are converted to summations:

$$p(\mathbf{s} = \mathbf{s}_0|\mathbf{x}) = \sum_{\forall \mathbf{y}:\mathscr{C}(\mathbf{y})=\mathbf{s}_0} p(\mathbf{y}|\mathbf{x}). \tag{14}$$

□

### Proof of Proposition 3.2

*Proof.* Without loss of generality, let $s_{0,1} = 1$, $s_{0,2} = 2$, $\ldots s_{0,c} = c$ and let $t_1, t_2$ through $t_{c-1}$ are the state transition times, i.e., $t_1$ is the transition from $s_{0,1} = 1$ to $s_{0,2} = 2$.

$$p(\mathbf{s} = \mathbf{s}_0|\mathbf{x}) \tag{15}$$

$$= \sum_{\forall \mathbf{y}:\mathscr{C}(\mathbf{y})=\mathbf{s}_0} p(\mathbf{y}|\mathbf{x}) \tag{16}$$

$$\propto \sum_{\forall \mathbf{y}:\mathscr{C}(\mathbf{y})=\mathbf{s}_0} \prod_{t=1}^{T} \Psi(y_t, y_{t-1}, x_t) \tag{17}$$

$$\begin{aligned} = \sum_{0<t_1<t_2<\ldots t_{c-1}\leq T} \Bigg\{ \Big( \prod_{t=1}^{t_1-1} \Psi(1,1,x_t)\Big) \Psi(2,1,x_{t_1}) \\ \Big( \prod_{t=t_1+1}^{t_2-1} \Psi(2,2,x_t)\Big) \Psi(3,2,x_{t_2}) \\ \Big( \prod_{t=t_2+1}^{t_3-1} \Psi(3,3,x_t)\Big) \Psi(4,3,x_{t_3})... \\ \Psi(c,c-1,x_{t_{c-1}}) \Big( \prod_{t=t_{c-1}+1}^{T} \Psi(c,c,x_t)\Big) \Bigg\} \end{aligned} \tag{18}$$

$$\begin{aligned} = \Psi(c,c-1,x_T) \Bigg\{ \sum_{0<t_1<t_2<\ldots t_{c-2}\leq T-1} \Big( \prod_{t=1}^{t_1-1} \Psi(1,1,x_t)\Big) \\ \Psi(2,1,x_{t_1}) \Big( \prod_{t=t_1+1}^{t_2-1} \Psi(2,2,x_t)\Big) \Psi(3,2,x_{t_2}) \cdots \\ \Psi(c-1,c-2,x_{t_{c-2}}) \Big( \prod_{t=t_{c-2}+1}^{T-1} \Psi(c-1,c-1,x_t)\Big) \Bigg\} \end{aligned} \tag{19}$$

[1] The proofs are given for one dimensional random variable $\mathbf{y}$, however, it is trivial to extend them to T-tuple variable $\mathbf{y}$.

$$+\Psi(c,c,x_T)\Big\{\sum_{0<t_1<t_2<\dots t_{c-1}\leq T-1}\Big(\prod_{t=1}^{t_1-1}\Psi(1,1,x_t)\Big) \tag{20}$$

$$\Psi(2,1,x_{t_1})\Big(\prod_{t=t_1+1}^{t_2-1}\Psi(2,2,x_t)\Big)\Psi(3,2,x_{t_2})\cdots$$
$$\Psi(c,c-1,x_{t_{c-1}})\Big(\prod_{t=t_{c-1}+1}^{T-1}\Psi(c,c,x_t)\Big)\Big\} \tag{21}$$

In equation 18 we rewrite equation 17 by using the distributive law. In equation 19, we divide the summation into two cases by only factoring out time $T$: (1) First part considers the case in which there is a transition from $c-1$ to $c$ at time $T$; (2) Second part considers no transition at time $T$, so transition from $c-1$ to $c$ was before $T$ and at time $T$ the previous state $c$ is repeated. Next, let's define the forward variable for the **s** domain as $\overline{\alpha}_T(c)$ as:

$$\overline{\alpha}_T(c)=\sum_{0<t_1<t_2<\dots t_{c-1}\leq T}\Big\{\Big(\prod_{t=1}^{t_1-1}\Psi(1,1,x_t)\Big)\Psi(2,1,x_{t_1})$$
$$\Big(\prod_{t=t_1+1}^{t_2-1}\Psi(2,2,x_t)\Big)\Psi(3,2,x_{t_2})$$
$$\Big(\prod_{t=t_2+1}^{t_3-1}\Psi(3,3,x_t)\Big)\Psi(4,3,x_{t_3})\cdots$$
$$\Psi(c,c-1,x_{t_{c-1}})\Big(\prod_{t=t_{c-1}+1}^{T}\Psi(c,c,x_t)\Big)\Big\} \tag{22}$$

Then it is obvious that two summations in equation 19 can be written in terms of these forward variables:

$$\overline{\alpha}_{T-1}(c-1)=\Big\{\sum_{0<t_1<t_2<\dots t_{c-2}\leq T-1}\Big(\prod_{t=1}^{t_1-1}\Psi(1,1,x_t)\Big)$$
$$\Psi(2,1,x_{t_1})\Big(\prod_{t=t_1+1}^{t_2-1}\Psi(2,2,x_t)\Big)$$
$$\Psi(3,2,x_{t_2})\cdots\Psi(c-1,c-2,x_{t_{c-2}})$$
$$\Big(\prod_{t=t_{c-2}+1}^{T-1}\Psi(c-1,c-1,x_t)\Big)\Big\} \tag{23}$$

$$\overline{\alpha}_{T-1}(c)=\Big\{\sum_{0<t_1<t_2<\dots t_{c-1}\leq T-1}\Big(\prod_{t=1}^{t_1-1}\Psi(1,1,x_t)\Big)$$
$$\Psi(2,1,x_{t_1})\Big(\prod_{t=t_1+1}^{t_2-1}\Psi(2,2,x_t)\Big)\Psi(3,2,x_{t_2})...$$
$$\Psi(c,c-1,x_{t_{c-1}})\Big(\prod_{t=t_{c-1}+1}^{T-1}\Psi(c,c,x_t)\Big)\Big\} \tag{24}$$

Finally, we reach to the recursion formula:

$$\overline{\alpha}_T(c)=\Psi(c,c-1,x_T)\overline{\alpha}_{T-1}(c-1)+\Psi(c,c,x_T)\overline{\alpha}_{T-1}(c)$$

This proof is valid for all lengths $c$. In other words, we can think of the signal from 1 to $c-1$ as our signal of interest and, at any such input, the recursion at time T can be written as

$$\overline{\alpha}_T(i)=\Psi(i,i-1,x_T)\overline{\alpha}_{T-1}(i-1)+\Psi(i,i,x_T)\overline{\alpha}_{T-1}(i)$$

Moreover, this can be generalized to arbitrary $t=1.....T$ as well by recursing back in T. Thus, the final form is

$$\overline{\alpha}_t(i)=\Psi(i,i-1,x_t)\overline{\alpha}_{t-1}(i-1)+\Psi(i,i,x_t)\overline{\alpha}_{t-1}(i)$$

□